\documentclass[10pt,twocolumn,letterpaper]{article}

 \usepackage{cvpr}              

\definecolor{cvprblue}{rgb}{0.21,0.49,0.74}
\usepackage[pagebackref,breaklinks,colorlinks,allcolors=cvprblue]{hyperref}

\usepackage{amsmath}
\usepackage[accsupp]{axessibility}  
\usepackage{amssymb}
\usepackage{utfsym}  
\usepackage{verbatim}
\usepackage{soul, color, xcolor}
\usepackage{multirow}
\usepackage{pifont}
\usepackage{colortbl}  
\usepackage{hyperref} 
\usepackage{xcolor}

\definecolor{revise}{HTML}{ED2240}   
\definecolor{gray}{HTML}{F2F2F2}
\definecolor{urlcolor}{HTML}{ED028C}

\definecolor{linkcolor}{HTML}{ED2240}  

\hypersetup{
    colorlinks=true,
    linkcolor=linkcolor,
    urlcolor=urlcolor
}

\usepackage[capitalize]{cleveref}
\crefname{section}{Sec.}{Secs.}
\Crefname{section}{Section}{Sections}
\Crefname{table}{Table}{Tables}
\crefname{table}{Tab.}{Tabs.}

\usepackage[sort&compress]{natbib}
\def\paperID{3446} 
\def\confName{CVPR}
\def\confYear{2025}

\title{URWKV: Unified RWKV Model with Multi-state Perspective for Low-light Image Restoration}

\author{
    Rui Xu\textsuperscript{\rm 1,\rm 2}, Yuzhen Niu\textsuperscript{\rm 1,\rm 2}\thanks{Corresponding author.}, Yuezhou Li\textsuperscript{\rm 1,\rm 2}, Huangbiao Xu\textsuperscript{\rm 1,\rm 2}, Wenxi Liu\textsuperscript{\rm 1,\rm 2}, Yuzhong Chen\textsuperscript{\rm 1,\rm 2} \\
    \textsuperscript{\rm 1}Fujian Key Laboratory of Network Computing and Intelligent Information Processing,\\ College of Computer and Data Science, Fuzhou University, Fuzhou 350108, China\\
\textsuperscript{\rm 2}Engineering Research Center of Big Data Intelligence, Ministry of Education, Fuzhou 350108, China\\ 
  {\tt\small \{xurui.ryan.chn, yuzhenniu, liyuezhou.cm, huangbiaoxu.chn\}@gmail.com} \\
  {\tt\small  wenxi.liu@hotmail.com, yzchen@fzu.edu.cn} \\
}

\begin{document}
\maketitle
\begin{abstract}
Existing low-light image enhancement (LLIE) and joint LLIE and deblurring (LLIE-deblur) models have made strides in addressing predefined degradations, yet they are often constrained by  dynamically coupled degradations. To address these challenges, we introduce a Unified Receptance Weighted Key Value (URWKV) model with multi-state perspective, enabling flexible and effective degradation restoration for low-light images. Specifically, we customize the core URWKV block to perceive and analyze complex degradations by leveraging multiple intra- and inter-stage states. First, inspired by the pupil mechanism in the human visual system, we propose Luminance-adaptive Normalization (LAN) that adjusts normalization parameters based on rich inter-stage states, allowing for adaptive, scene-aware luminance modulation. Second, we aggregate multiple intra-stage states through exponential moving average approach, effectively capturing subtle variations while mitigating information loss inherent in the single-state mechanism. To reduce the degradation effects commonly associated with conventional skip connections, we propose the State-aware Selective Fusion (SSF) module, which dynamically aligns and integrates multi-state features across encoder stages, selectively fusing contextual information. In comparison to state-of-the-art models, our URWKV model achieves superior performance on various benchmarks,  while requiring significantly fewer parameters and computational resources. Code is available at: \href{https://github.com/FZU-N/URWKV}{\texttt{https://github.com/FZU-N/URWKV}}.

\end{abstract}

\begin{figure}[!t]\centering
	\includegraphics[width=0.94\linewidth]{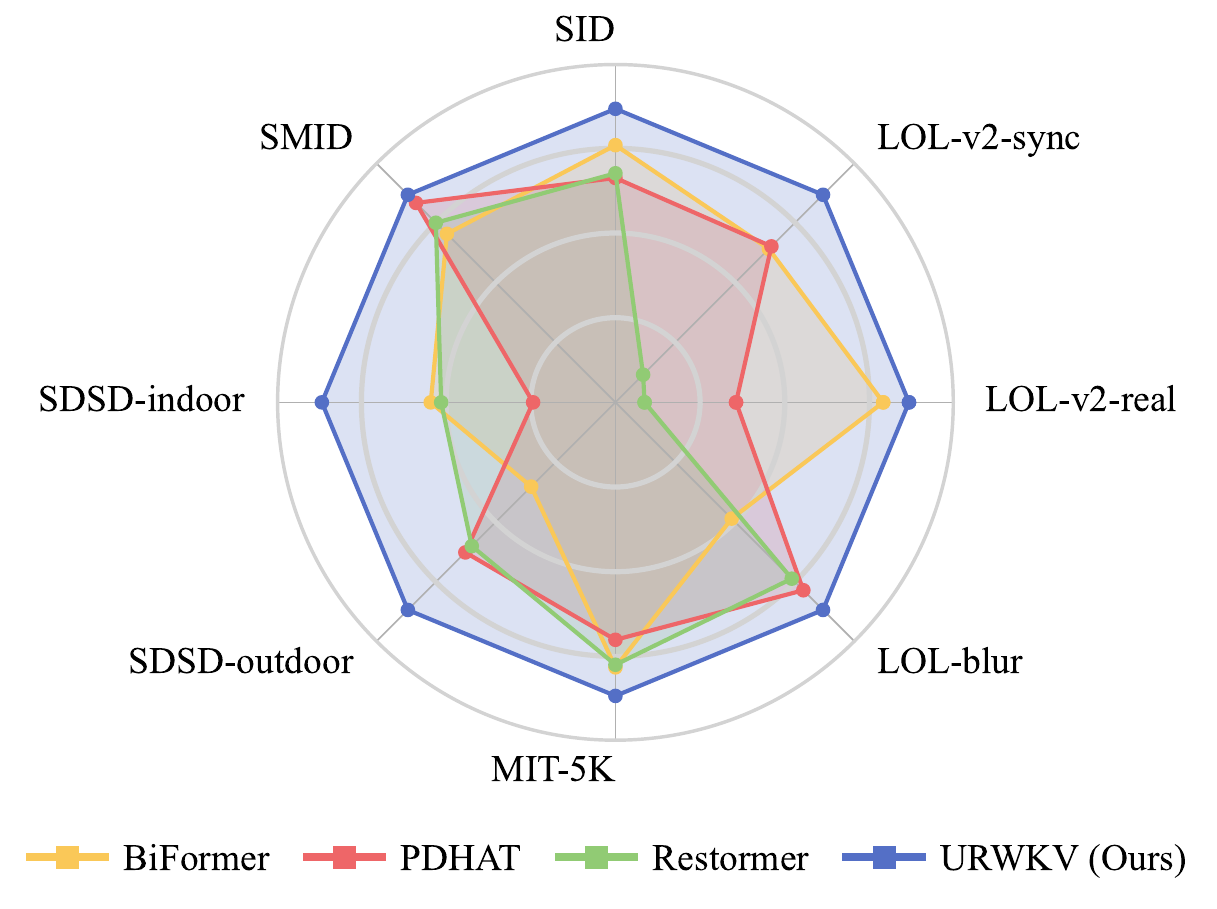}  
    \vspace{-2mm}
	\caption{Compared against existing solutions (including BiFormer \cite{BiFormer} for LLIE, PDHAT \cite{PDHAT} for joint LLIE and deblurring, and Restormer \cite{Restormer} as a unified model), our proposed URWKV achieves consistently superior PSNR performance across various  degradation scenarios. For better visualization, the maximum PSNR values across all datasets are normalized.
 } \label{introduction_fig}
\vspace{-6mm}
\end{figure}

\vspace{-4mm}
\section{Introduction}

Low-light environments present significant challenges in computer vision and pattern recognition, impacting applications such as nighttime surveillance \cite{visual_surveillance}, night photography \cite{night_photography}, and remote sensing \cite{remote_sensing}. In these scenarios, insufficient lighting often leads to severe image degradations, such as increased noise, loss of detail, reduced contrast, and color distortion. Furthermore, anothor degradation coupled in the low-light images, particularly motion blur induced by prolonged exposure times or camera shake, further compromise the integrity of the images. These challenges not only diminish  the visual quality of images but also impede subsequent image analysis and interpretation.

Low-light image enhancement (LLIE) models are foundational and widely researched solutions  for addressing  image degradations in low-light environments. Recent advancements have integrated   sophisticated architectures like CNNs \cite{KinD, RetinexNet}, Transformers \cite{Retinexformer, LLFormer}, and Mamba networks \cite{RetinexMamba}, enabling more effective handling of complex degradation patterns. To address specific challenges such as uneven brightness and spatially varying noise, several specialized methods \cite{SNR-Net, BiFormer} have been developed, enabling finer adjustments across different image regions. 
Despite these advancements, another typical coupled degradation, i.e., motion blur, that frequently occurs in low-light conditions, remains a significant challenge for current LLIE models.

The primary difficulty for addressing coupled degradations arises from the inherent trade-offs among brightness enhancement, noise reduction, and blur restoration. Traditional joint training strategy or the combination strategy of  off-the-shelf LLIE and deblurring models often fail to deliver optimal results \cite{zhou2022lednet, PDHAT}. To bridge this gap, Zhou \textit{et al.} \cite{zhou2022lednet} introduce a compact encoder-decoder architecture wherein the encoder focuses on low-light enhancement, while the decoder tackles blur restoration, effectively isolating the two tasks. Additionally, Li \textit{et al.} \cite{PDHAT} introduce a perceptual decoupling approach that incorporates auxiliary tasks to achieve task-specific representations. While these methods show efficacy in handling blur under low-light conditions, they are constrained by predefined degradation categories. As a result, their adaptability to the dynamically coupled degradations typically encountered in real-world scenarios is limited, where noise, blur, and other degradations may not occur in fixed, predictable combinations.

To flexibly address various degradations, the unified models \cite{MIRNet, IPT, Restormer, MambaIR} offer a natural attempt, emerging as a leading approach in image restoration. In particular, since IPT \cite{IPT} first demonstrates the potential of transformer architectures for low-level vision tasks, the development of transformer-based models \cite{Uformer, Restormer} has accelerated. However, significant challenges remain, particularly in addressing complex illumination and coupled degradation conditions. Many image restoration models, such as those based on transformers, lack adaptive mechanisms designed for low-light environments, often resulting in amplified degradation and the introduction of new artifacts. \textit{Specifically, two key issues persist: 1) the difficulty of exploiting  effective adaptive strategies for intricate illumination scenarios and handling coupled degradation factors, and 2) parameter inefficiency, which presents a critical bottleneck for practical deployment in downstream vision tasks. }


In this paper, we extend the existing unified encoder-decoder framework by integrating the recently proposed Receptance Weighted Key Value (RWKV) structure \cite{rwkv, vision-rwkv}, aiming to provide effective solutions for addressing dynamically coupled degradations in low-light conditions from a multi-state perspective. Specifically, we introduce both intra- and inter-stage states to perceive and analyze diverse degradations across multiple stages, enabling a more nuanced restoration throughout the processing pipeline. To achieve this, we first propose luminance-adaptive normalization (LAN), which dynamically adjusts normalization parameters based on rich inter-stage states, facilitating adaptive, scene-aware luminance modulation. Furthermore, we employ an exponential moving average (EMA) approach to aggregate multiple intra-stage states, effectively capturing subtle variations and mitigating the information loss inherent in single-state mechanisms. Instead of constraining our model to predefined degradation types, we exploit the relationships between multi-state representations to address spatially varying and coupled degradations, thus enabling flexible and dynamic restoration. Additionally, we introduce a state-aware selective fusion (SSF) module, which dynamically aligns and integrates multi-state features across encoder stages, selectively fusing contextual information to enhance restoration quality. As shown in Fig. \ref{introduction_fig}, our model consistently outperforms existing models across various degradation scenarios.

The main contributions of this paper are as follows: 
\begin{itemize}

\item We introduce the unified RWKV model with a novel multi-state perspective, enabling flexible and effective restoration of dynamically coupled degradations in low-light conditions.

\item We propose LAN to dynamically adjust normalization parameters based on inter-stage states, allowing for adaptive, scene-aware luminance modulation.

\item We aggregate multiple intra-stage states using an EMA-based approach to capture subtle degradation variations, while mitigating the long-range information loss.

\item We develop the SSF module, which dynamically aligns and integrates multi-state features across encoder stages, facilitating selective fusion of contextual information.

\item Our URWKV model outperforms existing models across diverse benchmarks, effectively handling dynamically coupled degradations with fewer parameters and lower computational cost.

\end{itemize}

\section{Related Work}
\subsection{Low-light Image Enhancement}

\begin{figure*}[!t]\centering
	\includegraphics[width=0.97\linewidth]{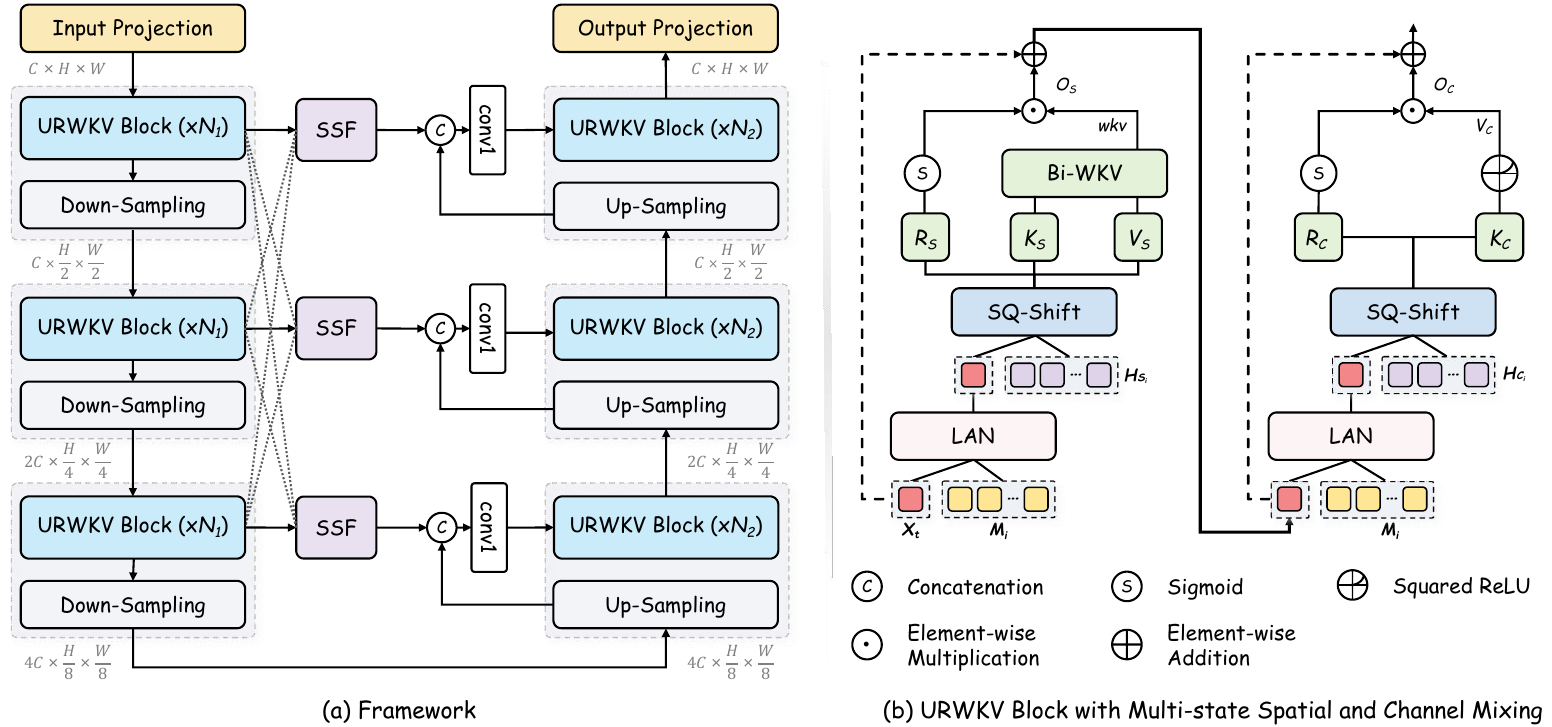}
	\caption{Overview of the URWKV model. The core URWKV block is integrated into an encoder-decoder framework, featuring a multi-state spatial mixing sub-block and a multi-state channel mixing sub-block. The liminance-adaptive normalization (LAN) incorporates  both the current input state $X_t$ and multiple inter-stage states $M_i$. Additionally, the multi-state quad-directional token shift (SQ-Shift) aggregates both current input state and multiple intra-stage states $H_i$ ($H_{S_i}$ for spatial mixing and $H_{C_i}$ for channel mixing, respectively). The state-aware selective fusion (SSF) module further aggregates rich contextual information across encoder stages. 
 } \label{Method_pipeline}
 \vspace{-2mm}
\end{figure*}

Compared to traditional methods relying on histogram equalization \cite{stark2000adaptive, abdullah2007dynamic, celik2011contextual} or Retinex-based enhancement \cite{fu2015probabilistic, li2018structure, jobson1997multiscale}, deep learning approaches have gained prominence in low-light image enhancement for their ability to handle complex degradations.  Retinex-based models, in particular, have evolved through the integration of advanced architectures such as CNNs \cite{KinD, RetinexNet}, Transformers \cite{Retinexformer}, and Mamba networks \cite{RetinexMamba}, enabling more precise illumination-reflectance decomposition. Leveraging the long-range dependency modeling capabilities of Transformers, several methods \cite{LLFormer, BiFormer} have introduced modifications to overcome the limitations of Transformers in low-light scenarios, achieving improved computational efficiency and enhanced local modeling capabilities. In addition, recognizing that noise characteristics vary spatially, Xu \textit{et al.} \cite{SNR-Net} propose to separate signal from noise during the enhancement process, enabling more effective brightening while minimizing noise amplification. Furthermore, frequency-domain approaches \cite{FourLLIE, UHDFour} also contribute to LLIE by decomposing features across distinct frequency bands, proving effective in preserving fine details and reducing noise. 

Considering blur as a frequent coupled degradation in low-light scenes, one that conventional LLIE models inadequately address, Zhou \textit{et al.} \cite{zhou2022lednet} and Li \textit{et al.} \cite{PDHAT} propose a specific sequential encoder-decoder architecture and a parallel heterogeneous decoupling framework, respectively. Although these LLIE-deblur models show effectiveness within certain constraints, further exploration is warranted to handle the dynamically coupled degradations typical in real-world scenarios.

\subsection{Unified Model for Image Restoration}

With the emergence of deep learning, unified approaches have driven substantial progress, enabling the handling of multiple restoration tasks within a single framework. Compared to CNN-based methods \cite{MIRNet, cui2024revitalizing}, which primarily focus on multi-scale feature extraction and local context enhancement, Transformer-based methods \cite{IPT, Restormer, liang2021swinir, dudhane2023burstormer} extend relational modeling capabilities by capturing long-range dependencies, leading to improved restoration quality across diverse degradation types. However, the application of Transformers in image restoration often entails high computational costs and memory requirements, which can hinder their efficiency. In addition, frequency-domain models \cite{zhou2023fourmer, cui2023image} provide another unified approach by decomposing images into multiple frequency bands, which enables focused noise reduction while preserving fine details across tasks. Recent innovations also utilize Mamba networks \cite{MambaIR} to achieve effective image restoration by efficiently balancing local and global feature representations. 

While current unified models perform effectively under normal lighting conditions, they encounter significant challenges in low-light environments, specifically involving complex lighting variations and coupled degradations.

\begin{figure*}[!t]\centering
	\includegraphics[width=0.96\linewidth]{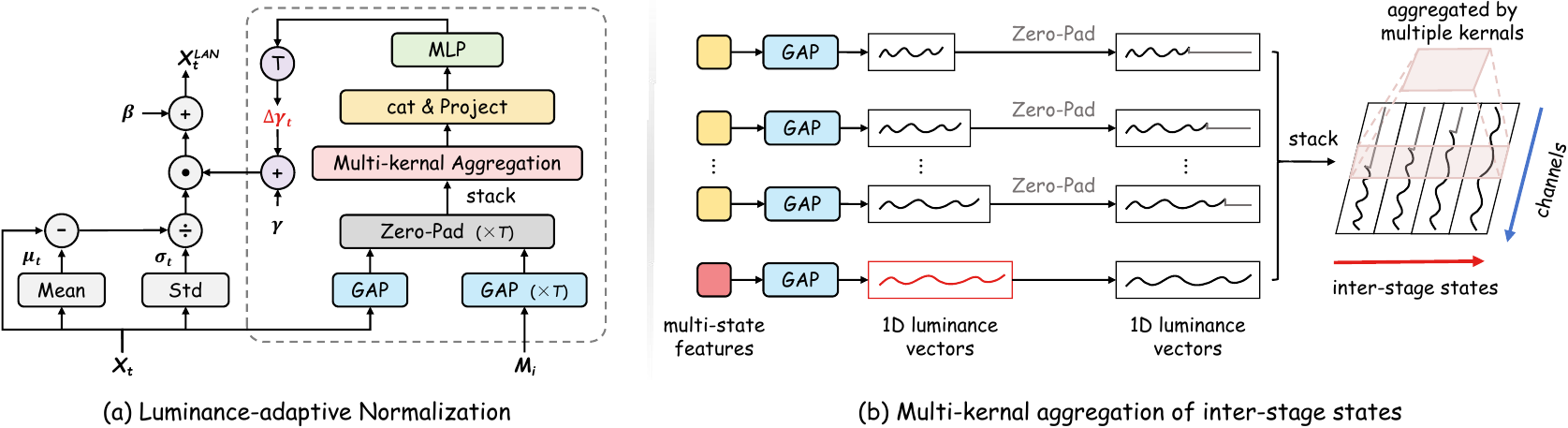}
	\caption{Illustration of luminance-adaptive normalization (LAN). LAN integrates inter-stage states $M_i$ throughout the restoration process, facilitating scene-aware luminance modulation. 
 } \label{Method_LAN}
 \vspace{-2mm}
\end{figure*}

\section{Method}
\subsection{Overview}

As illustrated in Fig. \ref{Method_pipeline}, the URWKV framework follows the typical encoder-decoder architecture, with both the encoder and decoder comprising three stages. In the encoder, each stage is composed of $N_1$ URWKV blocks followed by a down-sampling layer, while each stage in the decoder contains $N_2$ URWKV blocks and an up-sampling layer. Additionally, each down-sampling or up-sampling layer includes convolutional operations and sampling techniques to adjust either the channel or spatial dimensions.

To effectively handle dynamically coupled degradations, each URWKV block operates as an independent and progressive processing stage, with each stage's output (an implicit restored result, serving as a `state') encapsulating stage-specific contextual semantics and restoration cues (e.g., luminance or content) for subsequent stages.

Specifically, the URWKV block builds on the strengths of original RWKV model \cite{vision-rwkv}, while enhancing its capabilities through two tailored sub-blocks: a multi-state spatial mixing sub-block and a multi-state channel mixing sub-block. The former captures spatial dependencies across multiple intra-stage states to enrich spatial information, while the latter enhances feature representations through multi-state channel-wise interactions. Here, the intra-stage states $H_{S_i}$ or $H_{C_i}$ are updated following the SQ-Shift in each sub-block, respectively, where $i \in \{1, \dots, N_1\}$ (or $i \in \{1, \dots, N_2\}$).  To ensure efficient information transfer between encoder and decoder stages without introducing degradation, we incorporate the state-aware selective fusion (SSF) module between these stages. The SSF dynamically aggregates multi-state features across all encoder stages, selectively propagating rich contextual information into the decoder.

\subsection{URWKV with Multi-state Perspective}

Original RWKV models \cite{rwkv, vision-rwkv} employ a single-state token shift mechanism, where each token’s relationship relies solely on the current and preceding state. This design introduces two primary limitations: First, it struggles to retain long-range dependencies, causing early information to gradually fade during restoration. Second, it fails to effectively capture complex interdependencies among degraded features, as a single state is insufficient to represent intricate coupled degradations. To this end, we propose URWKV devised with the multi-state perspective. Specifically, the URWKV block tailors the luminance-adaptive normalization (LAN) by incorporating multiple inter-stage states to adapt URWKV to low-light scenarios. Additionally, it integrates multiple intra-stage states to facilitate more extensive interactions across long-range dependencies.


\textbf{Luminance-adaptive Normalization (LAN).} Luminance distortion is a major factor leading to other coupled degradations in low-light images. Inspired by the human eye’s ability to adaptively adjust pupil size in response to varying luminance \cite{lara2014changes, barrionuevo2023photoreceptor}, LAN leverages inter-stage states throughout the restoration process (comprising six stages) to enable scene-aware luminance modulation. This design overcomes limitation of LayerNorm widely used in existing models \cite{Uformer, Restormer}, which relies on fixed parameters $\gamma$ for scaling the learned representation.

As illustrated in Fig. \ref{Method_LAN}, given the current state input $X_t \in \mathbb{R}^{C_t \times H_t \times W_t}$ at stage $T$, we first extract 1D luminance vectors from both $X_t$ and each historical stage output $M_i \in \mathbb{R}^{C_i \times H_i \times W_i}$, using the global average pooling (GAP) operation. Here, each $M_i$ is obtained from the output of each stage’s final URWKV block. $t \in \{1, ..., N_1\}$ (or $t \in \{1, ..., N_2\}$) and $i \in \{1, \dots, T\text{-}1\}$.  The luminance vectors are then zero-padded to match the dimensionality of the vector with the maximum channel size $C_{max}$, while preserving the semantic integrity of luminance vector from each individual state. These zero-padded luminance vectors, denoted as $X^{1D}_{k} \in\mathbb{R}^{C_{max}}$, for  $k \in \{1,...,T\}$,  are then stacked along the state dimension. Next, a multi-kernel aggregation layer is applied to  analyze the stacked 2D luminance map along the channel dimension. Specifically, the aggregation layer uses 1D convolutional kernels of sizes $1 \times T$, $3 \times T$, and $5 \times T$ to adaptively capture both local and broader contextual illuminance variations across multiple states. The resulting luminance maps $X^{2D}_{r} \in\mathbb{R}^{T \times C_{max}}$ are expressed as:
\begin{equation}
    X^{2D}_{r} = \mathrm{Conv}^{1D}_{r}([X^{1D}_1, ..., X^{1D}_T]), r\in \{1,3,5\},
\end{equation}
where $\mathrm{Conv}^{1D}_{r}$ denotes  the set of 1D convolutional kernels, and $[\cdot,\cdot]$ denotes the stacking  operation.

Subsequently, the output of each convolutional kernel is concatenated along the channel dimension and projected through a $1 \times 1$ convolution, resulting in the aggregated feature map $X_{\text{agg}} \in\mathbb{R}^{C_t}$. A multi-layer perceptron (MLP) along with a tanh activation function are then employed to predict the luminance modulator $\Delta \gamma_{t}$. The luminance modulator is computed as:

\begin{equation}
\Delta \gamma_{t} = \mathrm{tanh}( \mathrm{MLP} ( X_{\text{agg}} ) ),
\end{equation}
where $\mathrm{tanh}(\cdot)$ represents  the tanh activation function.

Once $\Delta \gamma_{t}$ is predicted, the scaling parameter $\gamma$ can be updated as $\hat{\gamma_{t}}  = \gamma + \Delta \gamma_{t}$. Ultimately, the input feature $X_t$ at stage  $T$ can be adaptively normalized with LAN, expressed as:

\begin{equation}
X_t^{{LAN}} = \frac{ X_t - \mu_t }{\sigma_t} \odot \hat{\gamma_{t}} + \beta,
\end{equation}
where $\mu_t$ and $\sigma_t$ represent the mean and standard deviation of the input feature $X_t$, respectively. $\beta$ is a shift parameter.

\textbf{Multi-state Quad-directional Token Shift (SQ-Shift).} Original quad-directional token shift (Q-shift) \cite{vision-rwkv} effectively captures spatial dependencies among neighboring tokens. However, it neglects the degradation relations between states, making it difficult to maintain consistent degradation restoration across consecutive states. To address this limitation, we aggregate multi-state features to extend Q-Shift, allowing the model to capture long-range dependencies across multiple states while preserving local spatial interactions. Specifically, before applying Q-shift, we employ the exponential moving average (EMA) approach to aggregate the current state with all preceding states within the same stage, enabling richer interactions with historical information. For instance, given the state feature $X^{LAN}_{t} \in\mathbb{R}^{C \times H \times W}$ normalized by LAN, and multiple historical intra-stage state features $H_i$, $i \in \{1, ..., t-1\}$, we define the multi-state aggregation $\mathrm{MSA}$ as follows:
\begin{equation}
\mathrm{MSA}(X^{LAN}_{t}) = \alpha \odot X^{LAN}_{t} + (1 - \alpha) \odot \mathrm{MSA}(H_{t-1}),  
\end{equation}
where $\alpha$ is a decay factor that controls the weight between the current state and the previously aggregated state.

\textbf{Multi-State Spatial and Channel Mixing.} For each input feature in the multi-state spatial mixing sub-block or multi-state channel mixing sub-block, we first apply LAN and SQ-Shift, resulting in features $X_{s}$ for spatial mixing or $X_c$ for channel mixing. In the spatial mixing process, $X_s$ undergoes three parallel linear projections to generate the matrices for receptance $R_s$, key $K_s$, and value $V_s$. Following the computational approach of \cite{vision-rwkv}, we employ the bidirectional attention mechanism $\mathrm{Bi\text{-}WKV}$ to compute the global attention weights $wkv$ between $K_s$ and $V_s$, which are subsequently modulated by the sigmoid-activated receptance $R_s$ to control the output. The output feature $O_s$ is therefore represented as:
\begin{align}
      O_s = (\sigma(R_s)  \odot wkv)W_{O_s} ,
\end{align}
where $wkv = \mathrm{Bi\text{-}WKV}(K_{\mathrm{s}}, V_{\mathrm{s}})$. $\sigma$ represents the sigmoid function. $W_{O_s}$ is learnable parameter of projection layer.

For channel mixing, $X_c$ undergoes two parallel linear projections to derive the receptance matrix $R_c$ and key matrix $K_c$. Then, the output feature $O_c$ for channel mixing can be formulated as:
\begin{equation}
    O_c = (\sigma(R_c)  \odot \mathrm{SquaredReLU}(K_c)W_{V_c}){W_{O_c}} ,
\end{equation}
where $W_{V_c}$ and $W_{O_c}$ are learnable parameters in the two projection layers.

\begin{figure}[!t]\centering
	\includegraphics[width=0.99\linewidth]{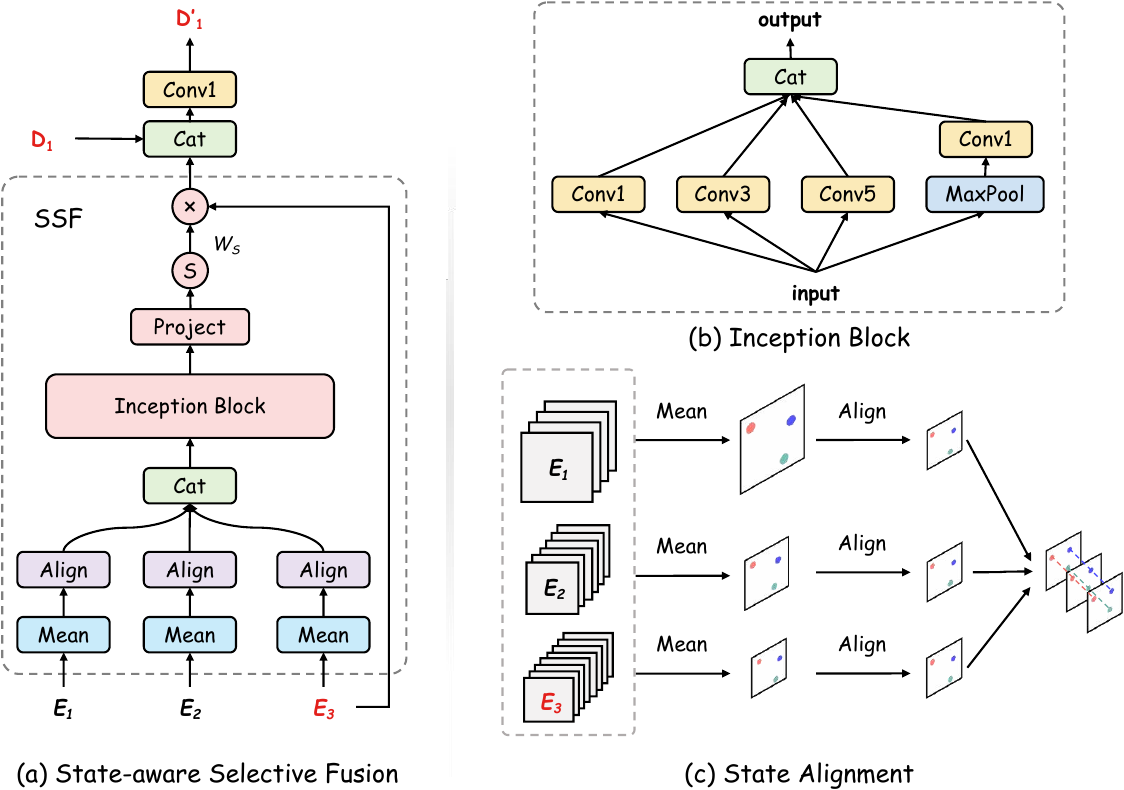}
     \vspace{-1mm}
	\caption{Illustration of the state-aware fusion (SSF) module.  
 } \label{Method_SSF}
 \vspace{-3mm}
\end{figure}

\begin{table*}[]
  \centering
\setlength{\tabcolsep}{4pt} 
  \renewcommand\arraystretch{1.3} 
  \caption{Quantitative comparison results on the LOL-v2-real, LOL-v2-syn, SID, SMID, SDSD-indoor, and SDSD-outdoor datasets. Joint LLIE and deblurring models are marked with $\textcolor{blue}{\tiny{\rm \dagger}}$, unified models are marked with $\textcolor{red}{\tiny{\rm \ast}}$, and the rest represent LLIE models. } 
  \vspace{-1mm}
   \resizebox{\linewidth}{!}{
\begin{tabular}{l|cc|cc|cc|cc|cc|cc| cc}
\toprule

\multirow{2}{*}{\textbf{Methods}}  & \multicolumn{2}{c|}{\textbf{LOL-v2-real}}  & \multicolumn{2}{c|}{\textbf{LOL-v2-syn}} & \multicolumn{2}{c|}{\textbf{SID}} & \multicolumn{2}{c|}{\textbf{SMID}} & \multicolumn{2}{c|}{\textbf{SDSD-indoor}} & \multicolumn{2}{c|}{\textbf{SDSD-outdoor}} & \multicolumn{2}{c}{\textbf{Complexity}} \\ \cline{2-15}

\textbf{}  & \textbf{PSNR} $\uparrow$ & \textbf{SSIM} $\uparrow$ & \textbf{PSNR} $\uparrow$  & \textbf{SSIM} $\uparrow$  & \textbf{PSNR} $\uparrow$  & \textbf{SSIM} $\uparrow$ & \textbf{PSNR} $\uparrow$  & \textbf{SSIM} $\uparrow$ & \textbf{PSNR} $\uparrow$  & \textbf{SSIM} $\uparrow$ & \textbf{PSNR} $\uparrow$  & \textbf{SSIM} $\uparrow$ & \textbf{FLOPs (G)}  & \textbf{Params (M)}  \\

\hline
\hline

KinD \cite{KinD}  & 14.74 & 0.641 & 13.29 & 0.578 & 18.02 & 0.583 & 22.18 & 0.634 & 21.95 & 0.672 &21.97 & 0.654 & 34.99 & 8.02 \\


EnlightenGAN \cite{EnlightenGAN} & 18.23 & 0.617 & 16.57 & 0.734 & 17.23 & 0.543 & 22.62 & 0.674 & 20.02 & 0.604 & 20.10 & 0.616 & 61.01 & 114.35\\

SNR-Net \cite{SNR-Net} & 21.48 & 0.849 & 24.14 & 0.927 & 21.35 & 0.550 & 28.52 & 0.807 & 26.13 & 0.815 & 19.22 & 0.657 & 26.35 & 4.01 \\

FourLLIE \cite{FourLLIE} & 22.35 & 0.847 & 24.65 & 0.919 & 18.42 & 0.513 & 25.64 & 0.750 & 24.74 & 0.826 & 24.67 & 0.787 & -& 0.12 \\

UHDFour \cite{UHDFour}  & 21.79 & 0.854 & 23.60 & 0.913 & 22.74 & 0.651& 28.18 & 0.809 & 29.13 & 0.895  & 23.28 & 0.771 & 57.42 & 15.90 \\

LLFormer \cite{LLFormer}  & 21.63 & 0.808 & 24.13 & 0.908 & 22.83 & 0.656 & 28.42 & 0.794 & 26.82 & 0.825 & 29.18 & 0.841 & 22.52 & 24.52 \\

Retinexformer \cite{Retinexformer}  & \textcolor{blue}{{22.79}} & 0.839 & 25.67 & 0.928 & \textcolor{red}{{24.44}} & \textcolor{red}{{0.680}} & 29.15 & 0.815 & \textcolor{blue}{{29.78}} & 0.895 & \textcolor{blue}{{29.83}} & \textcolor{blue}{{0.878}} & 15.57 & 1.61 \\

BiFormer \cite{BiFormer}  & 22.67 & \textcolor{blue}{{0.863}} & 24.87 & 0.921 & 22.49 & 0.638 & 28.24 & 0.812  & 28.73 & \textcolor{blue}{{0.899}} & 26.13 & 0.841 & 5.93 & 0.83 \\


RetinexMamba \cite{RetinexMamba} & 22.45 & 0.843 & \textcolor{blue}{{25.88}} & 0.933 & 22.45 & 0.656 & 28.62 & 0.809 & 28.44 & 0.894 & 28.52 & 0.859 & 34.75 & 24.1 \\

\hline
\hline

LEDNet$\textcolor{blue}{^{\rm \dagger}}$ \cite{zhou2022lednet} & 19.02 & 0.835 & 24.79 & 0.931 & 21.47 & 0.638 & 28.42 & 0.807 & 27.29 & 0.876 & 26.66 & 0.850 & 38.57 & 7.41\\

PDHAT$\textcolor{blue}{^{\rm \dagger}}$ \cite{PDHAT} & 20.16 & 0.841 & 24.94 & 0.937 & 21.93 & 0.653 & \textcolor{blue}{{29.19}} & \textcolor{blue}{{0.817}} & 26.37 & 0.884 & 28.19 & 0.862 & 208.19 & 7.83 \\
\hline
\hline


Uformer$\textcolor{red}{^{\rm \ast}}$ \cite{Uformer} & 18.82 & 0.771 & 19.66 & 0.871 & 18.54 & 0.577 & 27.20 & 0.792 & 23.17 & 0.859 & 23.85 & 0.748 & 12.00 & 5.29 \\

MIRNet$\textcolor{red}{^{\rm \ast}}$ \cite{MIRNet}  & 22.68 & 0.828 & 25.05 & 0.923 & 21.36 & 0.632 & 26.21 & 0.769 & 28.64 & 0.888 & 28.99 & 0.869 & 785.00  & 31.76 \\

Restormer$\textcolor{red}{^{\rm \ast}}$ \cite{Restormer} & 18.60 & 0.789 & 21.41 & 0.831 & 22.01 & 0.645 & 28.58 & 0.806 & 28.49 & 0.892 & 27.99 & 0.868  & 140.99 & 26.11 \\

MambaIR$\textcolor{red}{^{\rm \ast}}$ \cite{MambaIR} & 20.45 & 0.845 & 25.65 & \textcolor{blue}{{0.939}} & 22.02 & 0.658 & 28.41 & 0.805 & 25.14 & 0.876 & 27.53 & 0.851 & 60.66 & 4.30 \\
\hline
\hline

\textbf{URWKV (Ours)$\textcolor{red}{^{\rm \ast}}$} &\textcolor{red}{{23.11}} & \textcolor{red}{{0.874}}& \textcolor{red}{{26.36}}& \textcolor{red}{{0.944}} & \textcolor{blue}{{23.11}} & \textcolor{blue}{{0.673}} & \textcolor{red}{{29.44}}  & \textcolor{red}{{0.826}} &\textcolor{red}{{31.24}} & \textcolor{red}{{0.911}} & \textcolor{red}{{29.99}} & \textcolor{red}{{0.887}} & 18.34& 2.25 \\
\bottomrule
\end{tabular}
}
 \label{lowlight_table}%
 \vspace{-3mm}
\end{table*}

\subsection{State-aware Selective Fusion (SSF)} 


Exisiting models typically reintroduce early-stage features using navie skip connections (e.g., concatenation or addition). However, they are prone to two major issues, particularly under low-light conditions: (1) the propagation of noise and irrelevant information from the encoder to the decoder \cite{BiFormer}, and (2) the semantic gap between stages \cite{wang2022uctransnet}, which hinders effective feature fusion. 

Rather than adopting the navie or parameter-heavy approaches \cite{Uformer,Restormer,wang2022uctransnet}, we propose the lightweight and effective state-aware selective fusion (SSF) module. As illustrated in Fig. \ref{Method_SSF}, the SSF module focuses on capturing the degradation restoration relationships between multi-state features to predict the spatial guiding weight, which selectively refines the transfer of encoder features to the decoder. 
Given three encoder outputs $E_i \in \mathbb{R}^{C_i \times H_i \times W_i}$ obtained from $M_i$,  $i \in \{1, 2, 3\}$, we first apply a channel-wise mean operation to collapse the channel dimension, resulting in $E_i^{M} \in \mathbb{R}^{1 \times H_i \times W_i}$, which reduces inter-channel semantic interference.  Next, the adaptive alignment operation is performed on $E_i^{M}$ to adjust the spatial dimensions of each feature according to the target dimensions of the decoder at the current fusion stage. The aligned features $E_i^{A}$ are then stacked along the channel dimension to produce the concatenated feature map $E^{A} \in \mathbb{R}^{3 \times H_d \times W_d}$, where $H_d$ and $W_d$ are the height and width of the decoder's feature. This process is formally expressed as:
\begin{equation}
   E^{A} =  \mathrm{Stack}(\mathrm{AdaAlign}(\mathrm{Avg}(E_i))), i \in \{1, 2, 3\},
\end{equation}
where $\mathrm{Avg}$ denotes the channel-wise mean operation. $\mathrm{AdaAlign}$ combines both upsampling and downsampling to adaptively align multiple states with the target state. $\mathrm{Stack}$ is the operation that stacks the aligned features along the channel axis.

To facilitate more nuanced fusion of local details and broader context, we incorporate an inception block \cite{GoogLeNet} with multiple convolutional filters ($1 \times 1$, $3 \times 3$, and $5 \times 5$) to aggregate degradation patterns across various receptive fields. The aggregated feature $E_{agg}$ is then projected via a convolutional layer, followed by a sigmoid function to produce the spatial weight. Therefore, taking the first stage of the decoder as an example, the naive concatenation-based skip connection for the URWKV block in the decoder, following the SSF module, can be extended as follows:
\begin{equation}
    D'_1 = ([W_{s} \odot E_3,D_1]) W_p 
\end{equation}
where $W_{s}$ is the spatial weight predicted by the SSF module, $E_3$ is the encoder feature, and $D_1$ is the input feature to the first stage of the decoder. $\odot$ denotes element-wise multiplication, and $W_p$ represents a convolutional layer applied to refine the features. $[\cdot, \cdot]$ denote the concatenate operation.

\begin{figure*}[!t]\centering
	\includegraphics[width=0.99\linewidth]{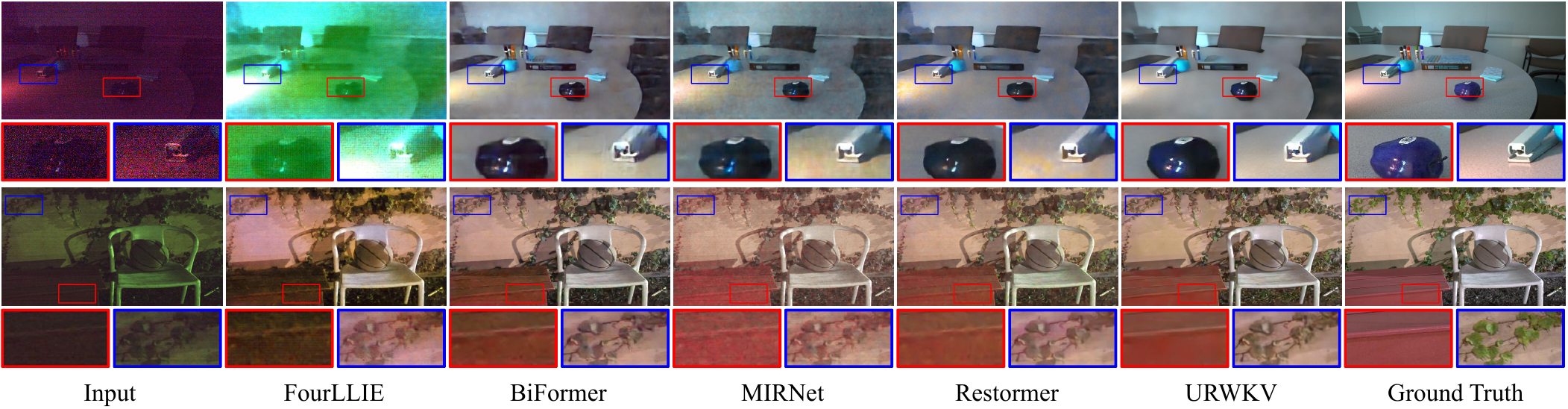}
    \vspace{-1mm}
	\caption{Visual comparison of state-of-the-art models on the SID dataset (top) and the SMID dataset (bottom). 
 } \label{Experiment_SID}
\vspace{-3mm}
\end{figure*}

\section{Experiments}
\subsection{Experimental Settings}

We implement our model in PyTorch on a workstation equipped with an NVIDIA Tesla A40 GPU. The number of URWKV blocks is set to $N_1=3$ and $N_2=2$, while the number of input channels is set to $C=32$. The decay factor $\alpha$ is set to 0.5. During training, we use a unified loss function for all datasets, as in \cite{BiFormer}, which combines L1 Loss, SSIM Loss, and Perceptual Loss. The Adam optimizer \cite{Adamkingma2014adamz} is employed with $\beta_1=0.9$ and $\beta_2=0.99$, and the learning rate is initialized at $2\times 10^{-4}$ and decays gradually to $10^{-6}$ using cosine annealing decay. Following \cite{LLFormer, BiFormer}, we apply augmentation techniques, such as flipping and rotation, to each image pair. During testing, our URWKV model processes input images of arbitrary shape directly, without the need for cropping or resizing. For evaluation, we employ commonly-used PSNR \cite{PSNR} and SSIM \cite{SSIM} metrics to calculate images in the RGB color space across  all datasets.

\subsection{Main Results}

To assess the performance of various models in scenarios involving dynamically coupled degradations, we compare the LLIE models, LLIE-deblur models, and unified models across 8 typical low-light benchmarks. 

\textbf{LOL. }  As shown in Table \ref{lowlight_table}, our URWKV demonstrates superior performance compared to LLIE models on LOL-v2-real \cite{LOL_v2}, outperforming Retinexformer by 0.32 dB in PSNR and 0.035 in SSIM, respectively.  Similarly, on LOL-v2-syn \cite{LOL_v2}, our model outperforms RetinexMamba by 0.48 dB in PSNR and 0.011 in SSIM. Notably, LLIE-deblur models \cite{zhou2022lednet, PDHAT} and other unified models \cite{MIRNet, Restormer, MambaIR} generally underperform on both datasets, with substantial performance gaps between them and our model.


\textbf{SID and SMID.}  As shown in Table \ref{lowlight_table}, our URWKV model delivers superior performance, closely trailing Retinexformer on SID \cite{SID_SMID}, while outperforming it by 0.29 dB in PSNR on SMID \cite{SID_SMID}. In contrast, LLIE-deblur models exhibit inconsistent performance across the datasets. For instance, while PDHAT ranks second on SMID, it significantly underperforms on SID, achieving a PSNR of just 21.93 dB, which is 1.18 dB lower than our model. Visual comparisons in Fig. \ref{Experiment_SID} also demonstrate the excellent denoising capability of our model when enhancing brightness.


\begin{figure}[!t]\centering
	\includegraphics[width=0.99\linewidth]{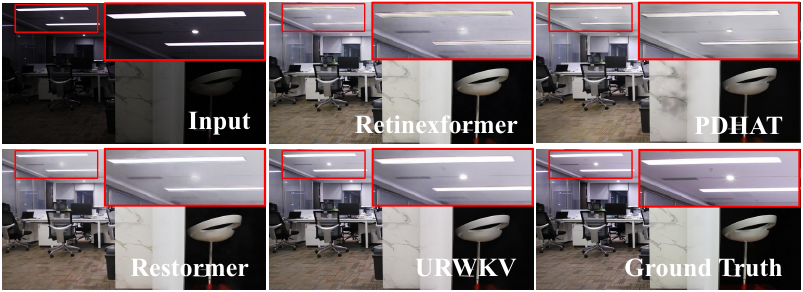}
    \vspace{-1mm}
	\caption{Visual comparisons on SDSD-indoor dataset. 
 } \label{Experiment_SDSD_indoor}
 \vspace{-4mm}
\end{figure}

\textbf{SDSD.}   Results depicted in Table \ref{lowlight_table} indicate that even some recent LLIE models \cite{FourLLIE, BiFormer} still encounter challenges on SDSD \cite{SDSD}. In contrast, our URWKV model delivers state-of-the-art performance across both subsets. Moreover, compared to unified models like Restormer, URWKV outperforms it by 2.75 dB in PSNR on SDSD-indoor  and by 2.00 dB in PSNR on SDSD-outdoor. As demonstrated in Fig. \ref{Experiment_SDSD_indoor}, our method preserves high-fidelity restoration in the scenario with strong light-dark contrast, whereas other models (e.g. PDHAT) generate unexpected artifacts.


\textbf{FiveK.}  The MIT-Adobe FiveK dataset \cite{MIT_5K} challenges models with the intricate task of color restoration in low-light and underexposed images. As reported in Table \ref{MIT_5K_table}, our model achieves the highest PSNR of 26.08 dB on FiveK dataset, surpassing LLFormer by 0.48 dB and Retinexformer by 1.50 dB. Furthermore, URWKV outperforms the unified model, Restormer, in SSIM by 0.007. These notable improvements underscore URWKV's proficiency in restoring both structural integrity and aesthetic quality. 



\begin{figure*}[!t]\centering
	\includegraphics[width=0.99\linewidth]{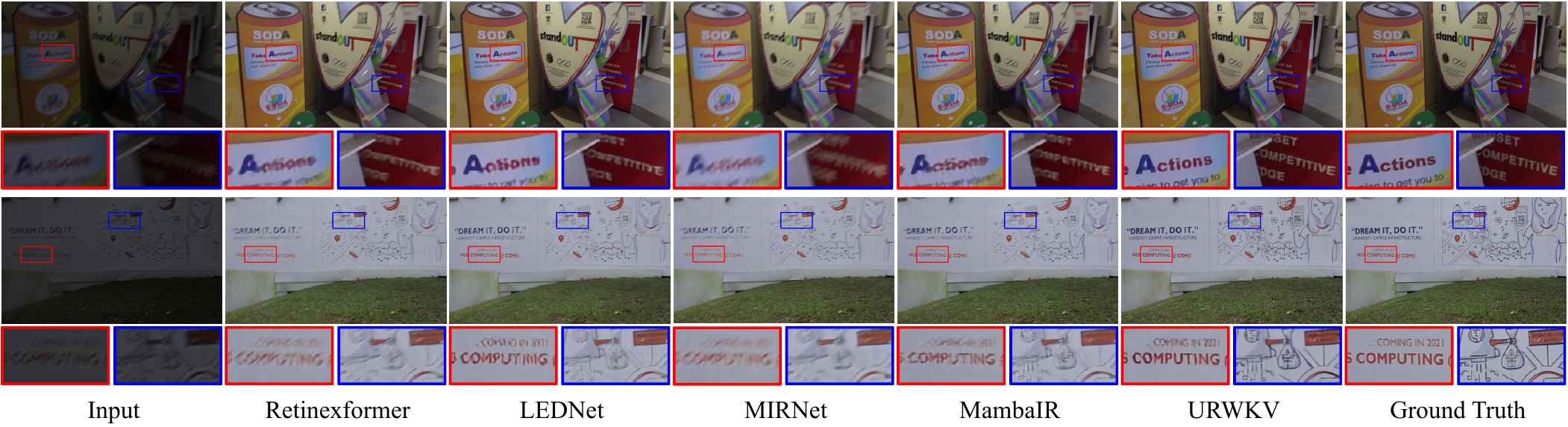}
    \vspace{-1mm}
	\caption{Visual comparison of state-of-the-art models on LOL-blur dataset. 
 } \label{Experiment_LOL_blur}
\vspace{-3mm}
\end{figure*}

\textbf{LOL-blur.} The LOL-blur \cite{zhou2022lednet} dataset couples varying degrees of low-light and motion blur, which has often been underrepresented in other LLIE datasets. As shown in Table \ref{LOL_blur_table}, conventional LLIE models typically struggle to address this coupled degradation, while unified models also fail to outperform specialized LLIE-deblur models.  In contrast, our URWKV model achieves the highest performance in this scenario, with a PSNR of 27.27 dB and SSIM of 0.890, outperforming the second-best LLIE-Deblur model PDHAT by 0.56 dB in PSNR and 0.011 in SSIM. Examples in Fig. \ref{Experiment_LOL_blur} vividly demonstrate the potential of our model in restoring coupled low-light and blur images.


\begin{table}[!t]
  \centering
  \setlength{\tabcolsep}{4pt} 
	\renewcommand\arraystretch{1.3}
  \caption{Experimental results on MIT-Adobe FiveK dataset.}
  \vspace{-2mm}
	\resizebox{\linewidth}{!}{
    \begin{tabular}{l|cccccc}
	  \toprule
	\multirow{2}{*}{\textbf{Methods}} & \textbf{EnlightenGAN}  & \textbf{RUAS} & \textbf{RetinexNet} & \textbf{SNR-Net} & \textbf{LLFormer} & \textbf{BiFormer} \\
	  & {\cite{EnlightenGAN}} & {\cite{RUAS}} & {\cite{RetinexNet}} & {\cite{SNR-Net}} & {\cite{LLFormer}}& {\cite{BiFormer}} \\
	\hline
	\textbf{PSNR $\uparrow$} & 13.26 & 20.91 & 16.32  & 24.39 & \textcolor{blue}{25.60} & 25.51 \\
	\textbf{SSIM $\uparrow$} & 0.745 & 0.863 & 0.782  & 0.895 & 0.917 & \textcolor{blue}{0.929} \\
	\hline
    \hline
	\multirow{2}{*}{\textbf{Methods}} & \textbf{Retinexformer} & \textbf{LEDNet} & \textbf{PDHAT} & \textbf{Restormer} & \textbf{MIRNet} & \textbf{URWKV} \\
	& {\cite{Retinexformer}} & {\cite{zhou2022lednet}} & {\cite{PDHAT}} & {\cite{Restormer}} & {\cite{MIRNet}} & \textbf{(Ours)} \\
	\hline
	\textbf{PSNR $\uparrow$} & 24.58 & 24.34 & 25.00 & 25.48 & 23.73 & \textcolor{red}{{26.08}} \\
	\textbf{SSIM $\uparrow$} & 0.886 & 0.915 & \textcolor{blue}{0.929} & \textcolor{blue}{0.929} & 0.895 & \textcolor{red}{{0.936}} \\
	\bottomrule
    \end{tabular}%
	}
 \vspace{-1mm}
  \label{MIT_5K_table}%
\end{table}%

\begin{table}[!t]
  \centering
    \setlength{\tabcolsep}{4pt} 
	\renewcommand\arraystretch{1.3}
  \caption{Experimental results on LOL-blur dataset.}
  \vspace{-2mm}
	\resizebox{\linewidth}{!}{
    \begin{tabular}{l|cccccc}
	  \toprule
	\multirow{2}{*}{\textbf{Methods}} & \textbf{FourLLIE} & \textbf{UHDFour} & \textbf{LLFormer} & \textbf{Retinexformer} & \textbf{BiFormer} & \textbf{GLARE} \\
	 & {\cite{FourLLIE}} & {\cite{UHDFour}} & {\cite{LLFormer}} & {\cite{Retinexformer}} & {\cite{BiFormer}}& {\cite{GLARE}} \\
	\hline
	\textbf{PSNR $\uparrow$}  & 19.81 & 25.34 & 24.55 & 25.25 & 24.67& 23.26 \\
	\textbf{SSIM $\uparrow$}  & 0.683 & 0.829 & 0.785 & 0.821 & 0.813 & 0.690\\
	\hline
    \hline
	\multirow{2}{*}{\textbf{Methods}} & \textbf{LEDNet} & \textbf{PDHAT} & \textbf{Restormer} & \textbf{MIRNet} & \textbf{MambaIR} & \textbf{URWKV} \\
	& {\cite{zhou2022lednet}} & {\cite{PDHAT}} & {\cite{Restormer}} & {\cite{MIRNet}} & {\cite{MambaIR}} & \textbf{(Ours)} \\
	\hline
	\textbf{PSNR $\uparrow$} & 26.06 & \textcolor{blue}{{26.71}} & 26.38 & 23.99 & 26.28 & \textcolor{red}{{27.27}} \\
	\textbf{SSIM $\uparrow$} & 0.846 & \textcolor{blue}{{0.879}} & 0.860 & 0.774 & 0.848 & \textcolor{red}{{0.890}} \\
	\bottomrule
    \end{tabular}%
	}
 \vspace{-4mm}
  \label{LOL_blur_table}%
\end{table}%

\subsection{Ablation Study}
We conduct several ablation studies on the real-world dataset (i.e., LOL-v2-real) to analyze the impact of various components in our model.
 
\begin{figure}[!t]\centering
	\includegraphics[width=0.99\linewidth]{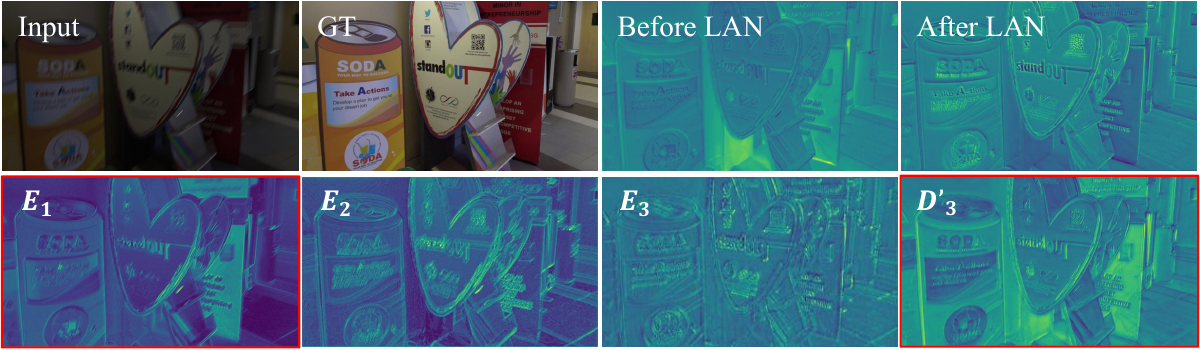}
     \vspace{-1mm}
	\caption{Top: LAN adaptively enhances the luminance, making details clearer. Bottom: SSF helps mitigate introducing degradation, obtaining fused feature ($D'_3$) based on $E_1$, $E_2$ and $E_3$.
 } \label{feature_transformation}
 \vspace{-5mm}
\end{figure}

\textbf{Effectiveness of LAN and SSF modules.} To validate the individual and combined contributions of the LAN and SSF modules, we evaluate four model variants. As shown in Table \ref{tab:LAN_and_SSF_modules}, models incorporating either LAN or SSF individually show a clear improvement over the baseline (first row in Table \ref{tab:LAN_and_SSF_modules}). Among these, the model with LAN alone achieves a more pronounced gain, with PSNR increasing by 1.38 dB over the baseline, highlighting the critical role of LAN in dynamically adapting to complex luminance variations. Moreover, our final model, which combines both LAN and SSF, incurs only a minor increase in parameters (from 1.64M to 2.25M) and a minimal rise in FLOPs (from 18.25G to 18.34G), yet delivers a significant performance boost compared to baseline. The feature transformation visualizations presented in Fig. \ref{feature_transformation} also validate the inherent advantages of LAN and SSF. 

\textbf{Multi-state Quad-directional Token Shift.} Our propose multi-state aggregation strategy enhances the original Q-shift, allowing for a more comprehensive understanding of degradation representation.  As shown in Table \ref{tab:token_shift_analysis}, our results indicate that single-state aggregation alone (Table \ref{tab:token_shift_analysis}a) achieves performance comparable to the model using Q-shift alone (Table \ref{tab:token_shift_analysis}c). In contrast, multi-state aggregation alone (Table \ref{tab:token_shift_analysis}b) consistently demonstrates superior performance, attaining a 0.13 dB higher PSNR. When combining single-state aggregation with Q-shift (Table \ref{tab:token_shift_analysis}d), the model shows an improvement in PSNR. However, it still falls short in SSIM by 0.003 compared to multi-state aggregation alone. Ultimately, our final model (Table \ref{tab:token_shift_analysis}e), which integrates both Q-shift and multi-state aggregation, demonstrates substantial advantages over all other variants.


\textbf{SSF vs. Navie Skip Connections.} 
The ablation results in Table \ref{tab:skipconnection} highlight the effectiveness of SSF compared to naive skip connections, including simple addition (Add) and concatenation (Cat). When using multi-state encoder features, noise from earlier stages and semantic gaps between states lead to performance degradation in both PSNR and SSIM. In contrast, our SSF module, based on state-aware fusion strategy, avoids directly merging multi-state features and  predicts a guiding weight to selectively fuse useful patterns, mitigating issues of noise propagation and semantic misalignment. Notably, the ``Multi-state+SSF"  (i.e., our URWKV) yields significant improvements in PSNR and SSIM over ``Single-state+Cat" without increasing the model’s parameter, demonstrating both enhanced performance and computational efficiency.

\begin{table}[!t]
\renewcommand\arraystretch{1.1}
\centering
\setlength{\tabcolsep}{10pt} 
\caption{Ablation study on  LAN and SSF modules.}
\vspace{-2mm}
\resizebox{0.9\linewidth}{!}{
\begin{tabular}{cc|cccc}
\toprule
\rowcolor{gray} 
\textbf{LAN} & \textbf{SSF} & \textbf{PSNR $\uparrow$} & \textbf{SSIM $\uparrow$} & \textbf{Params} & \textbf{FLOPs} \\ 
\midrule
{} & {} & 21.33 & 0.856 & 1.64M & 18.25G \\
 {}  & \checkmark  &   21.40   &  0.861       &    1.65M   &     18.29G   \\
\checkmark  & {} &   22.71   &  0.869      &   2.25M     &   18.30G    \\
\checkmark  & \checkmark &  23.11    &  0.874       &   2.25M     &   18.34G     \\ 
\bottomrule
\end{tabular}
}
\label{tab:LAN_and_SSF_modules}
\vspace{-1mm}
\end{table}

\begin{table}[!t]
\centering 
\setlength{\tabcolsep}{8pt} 
\caption{Ablation study on  multi-state quad-directional token shift.} 
\vspace{-2mm}
\resizebox{0.99\linewidth}{!}{ 
    \renewcommand{\arraystretch}{1.1} 
    \begin{tabular}{c|cc|c|cc}
        \toprule
        \rowcolor{gray} 
        \rowcolor{gray} 
        {\#} &  \textbf{Single-state} & \textbf{Multi-state} & \textbf{Q-shift} & \multicolumn{1}{c}{\textbf{PSNR $\uparrow$}} & \multicolumn{1}{c}{\textbf{SSIM $\uparrow$}} \\
        \hline
        \hline
        (a) & \checkmark & {} & {} & 22.12 & 0.861   \\
        (b)  & {} & \checkmark & {} & 22.25 & 0.870  \\
        \hline
        \hline
        (c)   & {} & {} & \checkmark & 22.19 & 0.862  \\
        (d)   & \checkmark & {} & \checkmark & 22.29 & 0.867   \\
        (e) & {} & \checkmark  & \checkmark  & 23.11 & 0.874   \\

        \bottomrule
    \end{tabular}
}
\label{tab:token_shift_analysis}
\vspace{-1mm}
\end{table}

\begin{table}[!t]
\renewcommand\arraystretch{1.1}
\centering
\caption{Impacts of SSF and navie skip connections (Skip Conn.) with different state numbers (State No.).}
\vspace{-2mm}
\resizebox{0.99\linewidth}{!}{
\begin{tabular}{cc|cccc}
\toprule
\rowcolor{gray} 
\textbf{State No.} & \textbf{Skip Conn.} & \textbf{PSNR $\uparrow$} & \textbf{SSIM $\uparrow$} & \textbf{Params} & \textbf{FLOPs} \\ 
\midrule
 Single-state & Add & 22.66 & 0.872 & 1.86M & 14.68G \\
 Single-state &  Cat  &   22.71   &  0.869      &   2.25M     &   18.30G    \\
 Multi-state &  Cat  &  21.83    &  0.859       &   2.50M    &   23.89G     \\ 
  Multi-state & SSF   &  23.11    &  0.874       &   2.25M     &   18.34G     \\ 
\bottomrule
\end{tabular}
}
\label{tab:skipconnection}
\vspace{-4mm}
\end{table}

\vspace{-2mm}
\section{Conclusion}

To address the shortcomings of existing models, we propose URWKV with multi-state perspective, enabling flexible and effective restoration across dynamically coupled degradation scenarios. Specifically, we introduce three key innovations: luminance-adaptive normalization (LAN) for adaptive, scene-aware luminance modulation; an exponential moving average approach to aggregate intra-stage states for spatial-level token-shift; and the state-aware selective fusion (SSF) module for dynamic alignment and integration of multi-state features across encoder stages. In comparison to existing models, our model demonstrates superior performance while requiring significantly fewer parameters and computational resources.

\noindent\textbf{Acknowledgments.} This work was supported in part by the National Natural Science Foundation of China under Grant 62471142, in part by the Natural Science Foundation of Fujian Province, China under Grants 2023J01067, in part by Industry-Academy Cooperation Project under Grant 2024H6006, in part by the Collaborative Innovation Platform Project of Fuzhou City under Grant 2023-P-002, in part by the Key Technology Innovation Project for Focused Research and Industrialization in the Software Industry of Fujian Province, and in part by the Key Research and Industrialization Project of Technological Innovation in Fujian Province under Grant 2024XQ002.
 

{
    \small
    \bibliography{main}
}


\end{document}